\documentclass[10pt,twocolumn,letterpaper]{article}

\usepackage[final]{cvpr}      % To produce the REVIEW version
\definecolor{cvprblue}{rgb}{0.21,0.49,0.74}
\usepackage[pagebackref,breaklinks,colorlinks,allcolors=cvprblue]{hyperref}

\usepackage{amsmath}
\usepackage{amssymb}
\usepackage{makecell}
\usepackage{multirow}
\usepackage{csquotes}
\usepackage{svg}

\usepackage[table]{xcolor} % 支持表格着色功能
\usepackage{colortbl} % 提供表格颜色支持
\usepackage{booktabs}
\usepackage[utf8]{inputenc}
\usepackage{etoolbox}
\usepackage{graphicx} % 可选，CVPR论文常用
\def\confName{NNNN}
\def\confYear{0000}

\title{ReGLA: Efficient Receptive-Field Modeling with Gated Linear Attention Network}

\author{
    Junzhou Li\textsuperscript{\rm 1$\ast$},
    Manqi Zhao\textsuperscript{\rm 2},
    Yilin Gao\textsuperscript{\rm 3},
    Zhiheng Yu\textsuperscript{\rm 2},
    Yin Li\textsuperscript{\rm 2},
    Dongsheng Jiang\textsuperscript{\rm 2},
    Li Xiao\textsuperscript{\rm 1}\\[6pt]
    \textsuperscript{\rm 1}University of Science and Technology of China\\
    \textsuperscript{\rm 2}Huawei Technologies Co., Ltd.\\
    \textsuperscript{\rm 3}Shanghai University\\
    ljz0824@mail.ustc.edu.cn,\quad
    933475@gmail.com,\quad
    gaoyilin@shu.edu.cn,\quad \\
    yuzhiheng8@huawei.com,\quad
    sherrylee90@163.com,\quad
    dongsheng\_jiang@outlook.com,\quad \\
    xiaoli11@ustc.edu.cn\\[6pt]
    \textsuperscript{$\ast$}Work done during an internship at Huawei\\
    Both Junzhou Li and Manqi Zhao are co-first authors. \\
    Both Li Xiao and Dongsheng Jiang are corresponding authors.
}

\begin{document}
\maketitle

\begin{abstract}
Balancing accuracy and latency on high-resolution images is a critical challenge for lightweight models, particularly for Transformer-based architectures that often suffer from excessive latency. To address this issue, we introduce \textbf{ReGLA}, a series of lightweight hybrid networks, which integrates efficient convolutions for local feature extraction with ReLU-based  gated linear attention for global modeling. The design incorporates three key innovations: the Efficient Large Receptive Field (ELRF) module for enhancing convolutional efficiency while preserving a large receptive field; the ReLU Gated Modulated Attention (RGMA) module for maintaining linear complexity while enhancing local feature representation; and a multi-teacher distillation strategy to boost performance on downstream tasks. 
Extensive experiments validate the superiority of ReGLA; particularly the ReGLA-M achieves \textbf{80.85\%} Top-1 accuracy on ImageNet-1K at $224px$,  with only \textbf{4.98 ms} latency at $512px$.
Furthermore, ReGLA outperforms similarly scaled iFormer models in downstream tasks, achieving gains of \textbf{3.1\%} AP on COCO object detection and \textbf{3.6\%} mIoU on ADE20K semantic segmentation, establishing it as a state-of-the-art solution for high-resolution visual applications.
\end{abstract}

\begin{figure}[t]
\centering
\includegraphics[width=0.8\columnwidth]{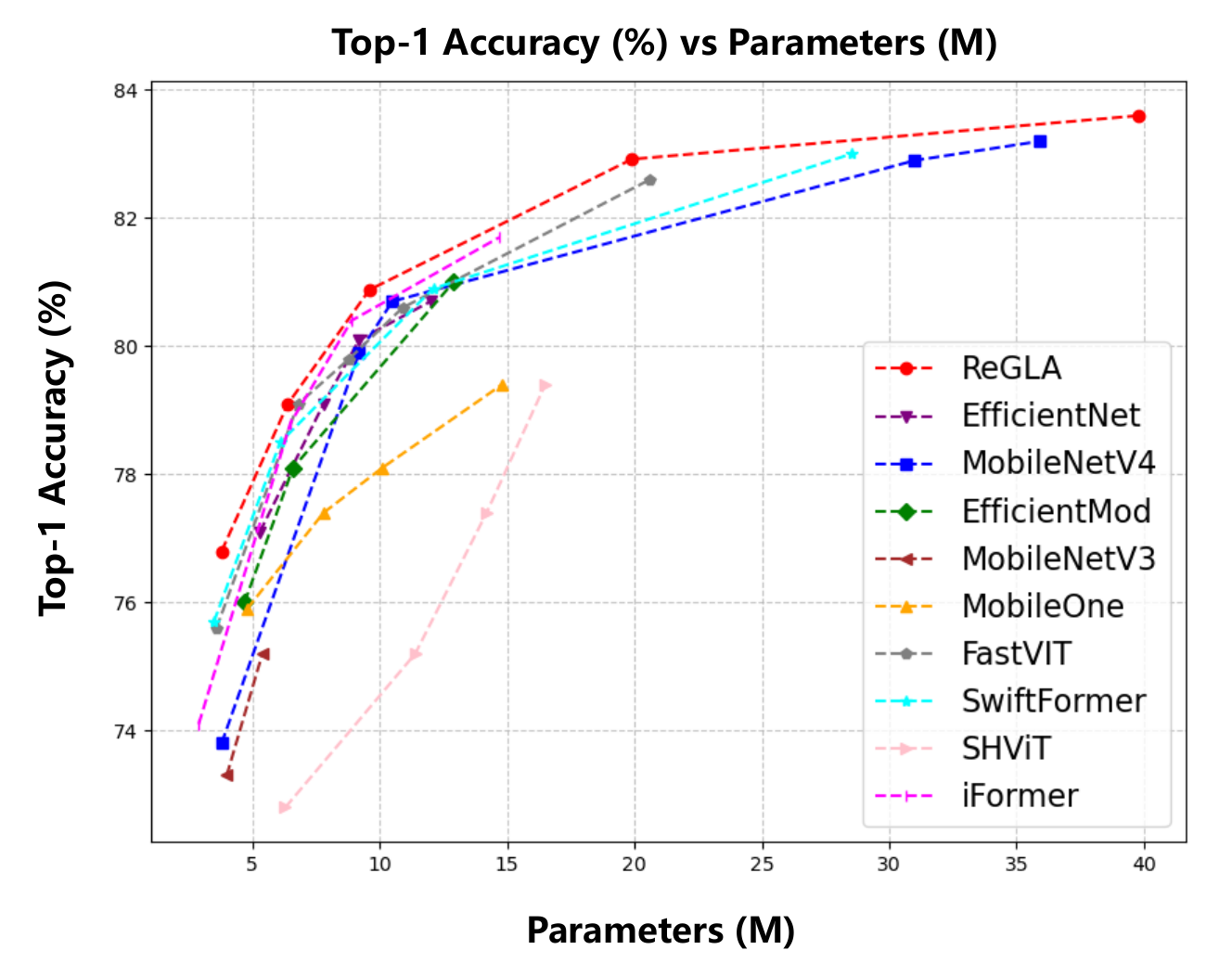} % Reduce the figure size so that it is slightly narrower than the column. Don't use precise values for figure width.This setup will avoid overfull boxes.
\caption{Comparison of Top-1 accuracy (\%) against parameter size (M) across various lightweight vision models. ReGLA consistently achieves competitive accuracy with similar parameters.}
\label{fig1}
\end{figure}

\section{Introduction}
\label{sec:intro}
\begin{quote}
    \textit{``Simplicity is the ultimate sophistication."}
    \begin{flushright}
        -- \textit{Leonardo da Vinci} 
    \end{flushright}
\end{quote}

The remarkable success of Vision Transformers (ViTs)~\cite{dosovitskiy2020image, han2022survey} has redefined the landscape of visual representation learning. By leveraging self-attention mechanisms to capture long-range dependencies, ViTs have achieved state-of-the-art performance across a wide spectrum of computer vision tasks~\cite{zhou2017ade20k, deng2009imagenet}. However, this advancement comes at a steep computational cost: the quadratic complexity of softmax-based attention with respect to sequence length renders standard ViTs impractical for deployment on resource-constrained platforms, such as mobile devices, embedded systems, and battery-powered IoT sensors.

This tension between model capability and real-world deployability has sparked a surge of research into efficient attention mechanisms and lightweight architectures~\cite{mehta2021mobilevit, cai2023efficientvit}. While linearized attention variants~\cite{zhang2023rethinking, shen2021efficient} reduce computational overhead by approximating or replacing the softmax kernel, they often sacrifice modeling fidelity or introduce numerical instability during training. Concurrently, hybrid Convolutional Neural Network (CNN)-Transformer designs~\cite{zheng2025iformer} attempt to balance local feature extraction with global modeling, yet many still inherit the inefficiencies of conventional attention.

In this work, we present \textbf{ReGLA}, a lightweight hybrid CNN-Transformer architecture that balances efficiency, stability and expressiveness through two key design principles: \emph{replacing softmax with a learnable, gated linear activation and maintaining a large receptive field while enhancing focus on local information}. Specifically, we first propose \textbf{ReLU-Gated Modulated Attention} (RGMA) — a memory-efficient attention mechanism that replaces softmax normalization with a ReLU-based linear kernel and a lightweight gating unit, reducing time and memory complexity to linear scaling in sequence length while ensuring stable, hardware-friendly gradients during training and inference. Additionally, \textbf{Efficient Large Receptive Field} (ELRF) is introduced to integrate two small separable depthwise convolutions interleaved with Feed-Forward Network (FFN) to achieve a larger receptive field and enhance local information, which is critical for high-resolution image processing.

Extensive experiments on ImageNet classification, COCO object detection, and ADE20K semantic segmentation show that ReGLA delivers superior accuracy with significantly reduced FLOPs, parameter count, and latency on mobile CPUs compared to recent efficient lightweight models (see Fig.~\ref{fig1}). Furthermore, multi-teacher distillation leveraging diverse pretrained models (SAM2, DeiT-III, DINOv2, etc.) greatly enhances its generalization capabilities.

In summary, the contributions of our work are as follows:
\begin{itemize}[leftmargin=*]
    \item We introduce ReGLA, a highly efficient CNN-Transformer architecture tailored for mobile and edge deployment.Specifically, ReGLA includes ELRF to enhance local features representation, and RGMA (a softmax-free attention mechanism by combining ReLU-based linear attention with dynamic channel-wise gating) to improve representational capacity.

    \item We design a multi-teacher distillation strategy, which addresses key challenges in multi-teacher model selection, conflict resolution, and handling the heterogeneity between teacher model (plain ViT) and student model (pyramidal Network) architectures.

    \item We provide comprehensive empirical validation across multiple vision benchmarks and hardware platforms, establishing ReGLA as a practical and scalable solution for sustainable computer vision.
\end{itemize} 

In an era where the environmental and economic costs of large-scale AI are under increasing scrutiny, ReGLA embodies a shift toward ``less but better''—proving that sophisticated visual intelligence need not come at the expense of efficiency or accessibility.

\section{Related Work}
\textbf{Efficient Convolutional Networks}. Over the past decade, CNNs have emerged as the dominant solution for various vision tasks. The research on lightweight CNN models is receiving increasing attention, and many high-quality research results have been proposed. For example,
MobileNets~\cite{mobilenets} pioneered depthwise separable convolution, reducing the computational costs; 
MobileNetV2~\cite{sandler2018mobilenetv2} introduced the inverted residual bottleneck module, using linear bottleneck structures and shortcut connections to reduce parameters while maintaining accuracy. 
EfficientNet~\cite{tan2019efficientnet} built a dynamic feature interaction mechanism by merging hierarchical features from context modeling and feature projection, improving the model's ability to represent multiscale visual data;
EfficientNetV2~\cite{tan2021efficientnetv2} proposed training-aware neural architecture search and scaling methods, and incorporates improved progressive learning.

However, pure convolutional architectures face limitations in spatial modeling~\cite{panahi2020spatial}, as their local receptive fields struggle to effectively capture long-range information correlations. Moreover, the high computational complexity of large convolution kernels conflicts with the limited memory and computing power of mobile devices, posing challenges for efficient deployment. In our work, we employ two small separable depthwise convolutions interleaved with FFN to expand the receptive field and strengthen the focus on local information.

\begin{figure*}[t]
\centering
\includegraphics[width=0.9\textwidth]{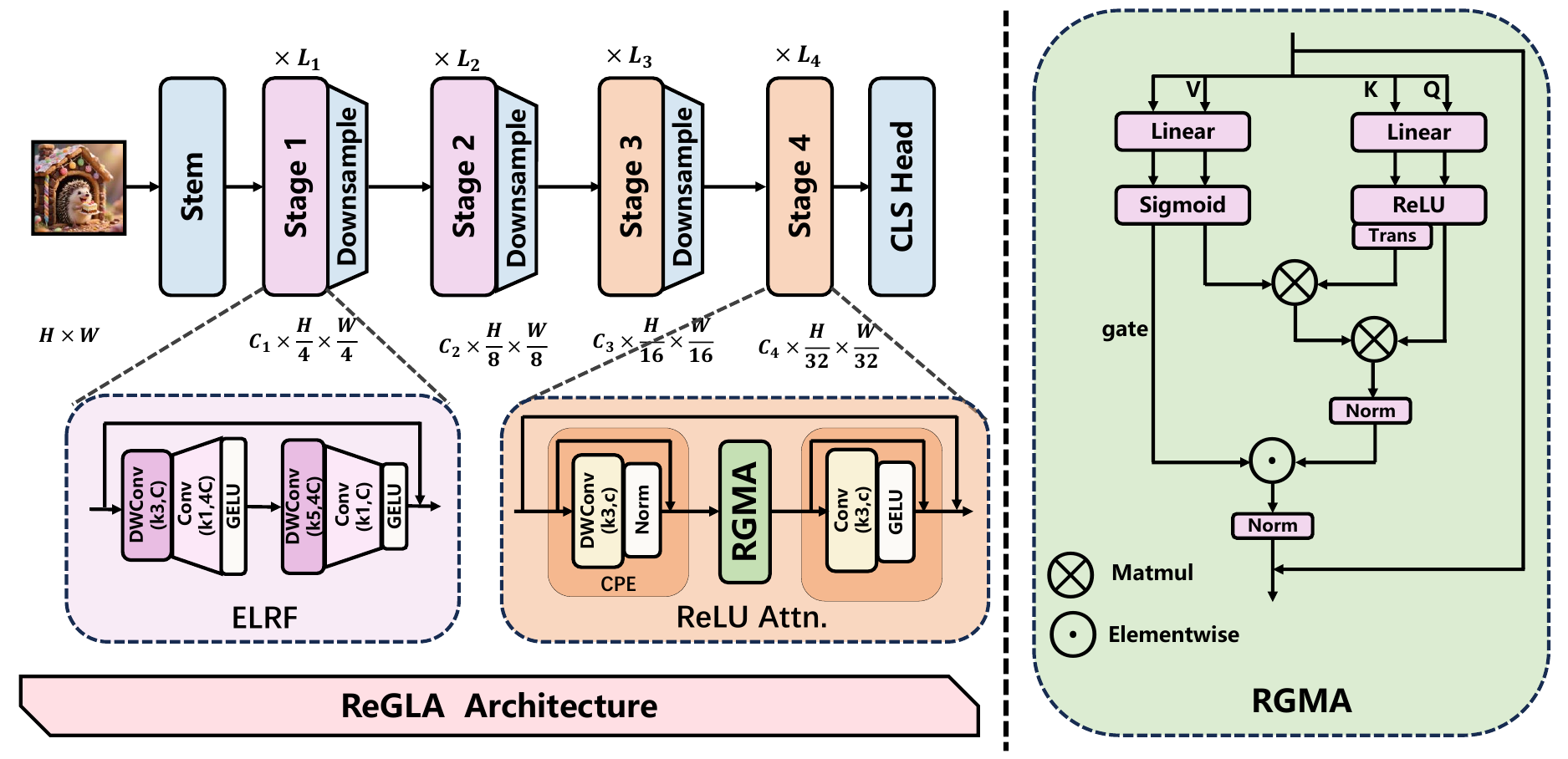} % Reduce the figure size so that it is slightly narrower than the column.
\caption{The architecture of ReGLA. In Stage 1 and Stage 2, we utilize ELRF to extract local information. In Stage 3 and Stage 4, RGMA is employed to focus on global information. DWConv denotes depthwise convolution. The CPE module is a depthwise convolution with residual connections. $L_i$ $(i=1,2,3,4)$ denotes the number of stages for each of the four stages, respectively. $C_i$ $(i=1,2,3,4)$ denotes the number of channels in each stage, respectively.}
\label{ReGLA}
\end{figure*}

\textbf{Efficient Hybrid Networks}. 
Recently, researchers have increasingly focused on developing efficient attention mechanisms for lightweight designs. A prominent direction is the integration of convolution and ViT to create compact models. Many lightweight ViT approaches adopt hybrid architectures, seeking to balance the global context modeling capabilities of transformers with the local feature extraction strengths of convolutional layers.

MobileViT~\cite{mehta2021mobilevit} combined MobileNetV2 with ViT to enhance the capabilities of lightweight vision models.  
MobileFormer~\cite{chen2022mobile} introduced a parallel architecture where MobileNet and ViT operated concurrently for improved efficiency.  
EfficientFormer~\cite{li2022efficientformer} and MobileNetV4~\cite{qin2024mobilenetv4} unified convolutional and Transformer operators into a shared search space, leveraging automated architecture search systems for optimal design.  
SHViT~\cite{yun2024shvit} addressed spatial redundancy by employing large-stride patch embedding, a 3-stage macro design, and single-head attention on partial channels to reduce multi-head redundancy. 
iFormer~\cite{zheng2025iformer} integrated optimized convolutions with efficient single-head modulated attention, achieving better mobile latency and accuracy. EfficientViT~\cite{cai2023efficientvit} proposed a hybrid architecture featuring multi-scale linear attention modules, optimizing hardware efficiency and balancing accuracy with linear complexity.

Unfortunately, the quadratic computational complexity and demands for computational resources pose critical challenges for deployment on edge devices. 
In our work, we develop RGMA, leveraging linear computational complexity while enhancing the local information extraction capability.

\section{Methods}
Our design prioritizes three principles: (1) linear computational complexity to accommodate high-resolution images, (2) maintaining a large receptive field while enhancing local information extraction capabilities, and (3) numerical stability during training.

\subsection{Preliminary: ReLU Linear Attention} 
ViTs rely on Multi-Head Self-Attention (MHSA) to model long-range dependencies. Given an input $\mathbf{X} \in \mathbb{R}^{N \times d}$, where $N$ is the number of tokens and $d_k$ is the embedding dimension, standard softmax attention~\cite{vaswani2017attention} with the Query (Q), Key (K), and Value (V) matrices compute
\begin{equation}
    \text{Attn}(\mathbf{Q}, \mathbf{K}, \mathbf{V}) = \text{Softmax}\left(\frac{\mathbf{Q}\mathbf{K}^\top}{\sqrt{d_k}}\right)\mathbf{V}.
    \label{eq:softmaxattn}
\end{equation}
Eq. (\ref{eq:softmaxattn}) yields quadratic computational complexity $\mathcal{O}(N^2d)$ due to pairwise token interactions. The softmax operation acts as a similarity measure, i.e.,
\begin{equation}
    \text{Softmax}\left(\frac{\mathbf{Q}_i \mathbf{K}_j^\top}{\sqrt{d_k}}\right) = \frac{Sim(\mathbf{Q}_i, \mathbf{K}_j)}{\sum_{j=1}^{N} 
    Sim(\mathbf{Q}_i, \mathbf{K}_j)},
    \label{eq:sim}
\end{equation}
where $Sim(\cdot)=\exp(\frac{\mathbf{Q}_i\mathbf{K}_j^\top}{\sqrt{d_k}})$.  In this work, we leverage the ReLU-based attention mechanism to achieve linear complexity~\cite{wortsman2023replacing}:
\begin{equation}
    \text{ReLUAttn}(\mathbf{Q}, \mathbf{K}, \mathbf{V}) = \frac{\text{ReLU}(\mathbf{Q}) \left( \text{ReLU}(\mathbf{K})^\top \mathbf{V} \right)}{\text{ReLU}(\mathbf{Q}) \left( \text{ReLU}(\mathbf{K})^\top \mathbf{1} \right)}.
    \label{eq:linearattn}
\end{equation}
This reduces complexity to $\mathcal{O}(Nd^2)$ by decoupling token interactions. While computationally efficient, the ReLU-based attention exhibits constrained local feature extraction capabilities compared to softmax formulations. Additionally, MHSA requires costly feature map reshaping operations that incur significant memory access overhead.

\subsection{ReLU Gated Modulated Attention (RGMA)} 
To overcome the limitations of standard ReLU attention in local feature extraction while preserving computational efficiency, we propose the RGMA module (see Fig.~\ref{ReGLA}). RGMA enhances local information capture through a bypass gated mechanism that selectively focuses attention, achieving optimal accuracy-latency balance for edge deployment.

Given a feature map $\mathbf{x} \in \mathbb{R}^{H \times W \times C}$ from preceding layers, where $C$, $H$, and $W$ respectively denote channel number, height and weight of the feature map. RGMA operates as
\begin{equation}
    \mathbf{X}_{o} = \mathcal{G}(\mathbf{x}) \odot \mathcal{C}(\mathbf{x}),
    \label{eq:RGMA}
\end{equation}
where $\mathcal{G}(\cdot)$ generates local feature weights via a convolutional gated mechanism, $\mathcal{C}(\cdot)$ computes global context through ReLU linear attention, and $\odot$ denotes element-wise multiplication. $\mathbf{X}_{o}$ is the output of the RGMA, which fuses fine-grained local patterns with efficient global representations.

In detail, the convolutional gated mechanism $\mathcal{G}(\cdot)$ is designed to extract local information representations, which is composed of a convolution and a nonlinear activation function,i.e.,
\begin{equation}
    \mathcal{G}(\mathbf{x}) = \sigma(\mathbf{W}_g \mathbf{x}),
    \label{eq:gate}
\end{equation}
where $\mathbf{W}_g$ denotes learnable convolutional weights, and $\sigma(\cdot)$ is the sigmoid activation producing spatial attention weights $[0,1]^{H \times W}$.

The ReLU linear attention $\mathcal{C}(\cdot)$ is designed to extract global information representations, which can generate memory-efficient global features by
\begin{equation}
    \mathcal{C}(\mathbf{x}) = \text{ReLUAttn}(\mathbf{W}_q \mathbf{x}, \mathbf{W}_k \mathbf{x}, \mathbf{W}_v \mathbf{x}),
    \label{eq:ctx}
\end{equation}
where $\mathbf{W}_q, \mathbf{W}_k, \mathbf{W}_v$ are projection weights. The output feature is processed through a reshaping operation to ensure compatibility in dimensions with $\mathcal{G}(\mathbf{x})$, thereby enabling element-wise fusion.

\subsection{Efficient Large Receptive Field Modeling (ELRF)} 
Early stages require expansive receptive fields to effectively capture spatial relationships in high-resolution inputs. 
To focus further on local information while maintaining a large receptive field, reducing parameter count and improving computational efficiency, we develop an efficient large receptive field design, called ELRF (see Fig.~\ref{ReGLA}). 

Building on the ExtraDW concept from MobileNetV4~\cite{qin2024mobilenetv4}, we strategically decompose large kernel operations into sequential depthwise convolutions. Our ELRF replaces traditional $7\times7$ depthwise convolutions with cascaded $3\times3$ and $5\times5$ depthwise operations interleaved with FFN layers. As illustrated in Fig.~\ref{ReGLA}, the process begins with a $3\times3$ depthwise convolution, followed by feature processing through an FFN, and concludes with a $5\times5$ depthwise convolution integrated within the FFN. This staged approach achieves the equivalent $7\times7$ receptive field while optimizing computational efficiency.

The decomposition ensures expansive spatial coverage and focuses further on local information, which is essential for early stages of high-resolution inputs, while substantially reducing computational complexity. By interleaving depthwise convolution operations with FFN layers, memory access bottlenecks are minimized, and the exclusive use of depthwise convolutions enhances hardware efficiency. This design effectively preserves the advantages of large-receptive-field modeling while addressing latency and efficiency constraints for mobile deployments.

\subsection{Network Architecture}
The ReGLA architecture comprises a stem block followed by four hierarchical stages, as illustrated in Fig.~\ref{ReGLA}. Given an input image with resolution $H \times W \times 3$, the stem performs initial feature extraction and spatial downsampling, outputting feature maps at $\frac{H}{4} \times \frac{W}{4} \times C_1$ resolution. 

Four subsequent stages progressively reduce spatial resolution while expanding channel capacity: Stage 1 ($\frac{H}{4} \times \frac{W}{4} \times C_1$), Stage 2 ($\frac{H}{8} \times \frac{W}{8} \times C_2$), Stage 3 ($\frac{H}{16} \times \frac{W}{16} \times C_3$), and Stage 4 ($\frac{H}{32} \times \frac{W}{32} \times C_4$). Spatial downsampling between stages is achieved through strided convolutions. Each stage contains $L_i$ identical processing blocks, where $i \in \{1,2,3,4\}$ denotes stage level, as shown in Fig.~\ref{ReGLA}. Model variants are constructed by scaling channel dimensions $C_i$ and block counts $L_i$, with full specifications in Table~\ref{ReGLA-tsmlx}.

\begin{table}[htbp]
\centering
\caption{Network Configurations of ReGLA Variants. The Blocks and Channels columns indicate the numbers of modules and channels for each stage, respectively. Detailed model architecture is provided in the supplementary materials.}
\resizebox{0.9\columnwidth}{!}{
\begin{tabular}{c|cc|c}
\toprule[1.2pt]
    Model       & Blocks (L)     & Channels (C)               &Param. (M)            \\  
\midrule[0.8pt]
    T           & [2, 2, 16, 6]    & [32, 64, 128, 256]        & 3.8                 \\
    S           & [2, 2, 19, 6]    & [32, 64, 160, 320]        & 6.4                 \\
    M           & [2, 2, 22, 6]    & [48, 96, 192, 384]        & 9.6                 \\      
    L           & [3, 3, 31, 9]    & [64, 128, 256, 448]       & 19.9                \\
    X           & [4, 4, 43, 12]   & [64, 128, 336, 512]       & 39.8                \\ 
\bottomrule[1.2pt]
\end{tabular}
}
\label{ReGLA-tsmlx}
\end{table}

\begin{table*}[!t]
\centering
\caption{\textbf{Classification results on ImageNet-1K.} The table is divided into five parts from top to bottom, presenting performance comparisons of different model sizes with magnitudes: 3M, 6M, 10M, 20M, and 40M. Our models achieve the state-of-the-art performance in different magnitudes. The Latency is measured by using coreML Tools on iPhone16 Pro of iOS 18.5, and '-' indicates that the Latency cannot be measured in Mobile here. We provide a more comprehensive comparison in the supplementary materials.}

\begin{tabular}{c|cc|c|ccc}
\toprule[1.2pt]
\multicolumn{1}{c|}{ \multirow{2}{*}{Model} } 
& \multicolumn{1}{c}{ \multirow{2}{*}{Param. (M)} }  
& \multicolumn{1}{c|}{ \multirow{2}{*}{GFlops (G)} }  
& \multicolumn{1}{c|}{Latency (ms)}
& \multicolumn{1}{c}{ \multirow{2}{*}{Reso.} } 
& \multicolumn{1}{c}{ \multirow{2}{*}{Epochs} } 
& \multicolumn{1}{c}{ \multirow{2}{*}{Top-1 (\%)} }                          \\

&
&
& \multicolumn{1}{c|}{224px/512px}
&
&
&  \\  

\midrule[0.8pt]

MobileNetV2~\cite{sandler2018mobilenetv2}        & 3.4       & 0.30     & 1.70 / 7.59             & 224   & 500    & 72.0                    \\
MNV3-Large~\cite{qian2021mobilenetv3}            & 4.0       & 0.153    & 1.77 / 6.72             & 224   & 600    & 73.3                    \\
MNV4-Conv-S~\cite{qin2024mobilenetv4}            & 3.8       & 0.184    & 1.39 / 5.82             & 224   & 500    & 73.8                    \\
iFormer-T~\cite{zheng2025iformer}                & 2.9       & 0.52     & 1.39 / 7.84             & 224   & 300    & 74.1                    \\
\rowcolor{gray!30}\textbf{ReGLA-T}               & 3.8       & 0.48     & 2.25 / \textbf{4.11}    & 224   & 300    & \textbf{76.4 (+2.3)}     \\ \midrule[0.8pt]

GhostNetV3~\cite{liu2024ghostnetv3}             & 6.1       & 0.17     & 1.97 / 12.27             & 224   & 600    & 77.1                   \\
RepViT-M1.0~\cite{wang2023repvit}               & 6.8       & 1.14     & 1.90 / 7.13             & 224   & 300    & 78.2                   \\
SwiftFormer-S~\cite{shaker2023swiftformer}      & 6.1       & 1.00     & 1.99 / 9.28             & 224   & 300    & 78.5                   \\
SHViT-S1~\cite{wang2023repvit}                  & 6.3       & 0.24     & 2.12 / 12.61             & 224   & 300    & 72.8                   \\
iFormer-S~\cite{zheng2025iformer}               & 6.5       & 1.08     & 1.90 / 14.05             & 224   & 300    & 78.8                   \\
\rowcolor{gray!30}\textbf{ReGLA-S}              & 6.4       & 1.04     & 2.76 / \textbf{4.36}     & 224   & 300    & \textbf{78.9 (+0.1)}    \\ \midrule[0.8pt]

EfficientViT-B1~\cite{cai2023efficientvit}      & 9.1       & 0.53     & 3.23 / 12.84             & 224   & 300    & 79.4                   \\
RepViT-M1.1~\cite{wang2023repvit}               & 8.2       & 1.37     & 2.27 / 8.30             & 224   & 300    & 79.4                   \\
MNV4-Conv-M~\cite{qin2024mobilenetv4}           & 9.7       & 1.70     & 2.35 / 8.80             & 256   & 500    & 79.9                   \\
MobileOne-S3~\cite{vasu2023mobileone}           & 10.07     & 1.90     & 2.53 / 14.02             & 224   & 300    & 80.0                   \\
GhostNetV31.6x~\cite{liu2024ghostnetv3}         & 12.3      & 0.40     & 2.83 / 15.13             & 224   & 600    & 80.4                   \\
iFormer-M~\cite{zheng2025iformer}               & 8.9       & 1.62     & 2.40 / 19.24             & 224   & 300    & 80.4                   \\
\rowcolor{gray!30}\textbf{ReGLA-M}              & 9.6       & 1.24     & 2.89 / \textbf{4.98}     & 224   & 300    & \textbf{80.85 (+0.45)}  \\ \midrule[0.8pt]

SHViT-S4~\cite{yun2024shvit}                    & 16.5      & 0.98     & 2.37 / 17.32             & 224   & 300    & 79.4                   \\
RepViT-M1.5~\cite{wang2023repvit}               & 14.6      & 2.34     & 2.75 / 13.55             & 224   & 300    & 81.2                   \\ 
iFormer-L~\cite{zheng2025iformer}               & 14.7      & 2.75     & 2.75 / 31.10             & 224   & 300    & 81.9                   \\
EfficientViT-B2~\cite{cai2023efficientvit}      & 24.3      & 1.59     & 4.52 / 19.12             & 224   & 300    & 82.1                   \\     
\rowcolor{gray!30}\textbf{ReGLA-L}              & 19.9      & 1.73     & 2.91 / \textbf{8.82}     & 224   & 300    & \textbf{82.9 (+0.8)}    \\ \midrule[0.8pt]

EfficientFormer-L3~\cite{li2022efficientformer} & 31.3      & 3.94     & 6.54 / 18.56                     & 224   & 300    & 82.4                  \\
MNV4-Conv-L~\cite{qin2024mobilenetv4}           & 31.0      & 2.17     & 4.16 / 16.36                     & 384   & 300    & 82.9                   \\
LowFormer-B1.5~\cite{nottebaum2025lowformer}    & 33.9      & 2.52     & 5.22 / 18.34                     & 224   & 300    & 81.2                 \\
EfficientViT-B3~\cite{cai2023efficientvit}      & 49.0      & 4.0      & 5.36 / 22.64                     & 224   & 300    & 83.4                   \\
MNV4-hybrid-L~\cite{qin2024mobilenetv4}         & 48.6      & 2.59     & -   /  -                         & 384   & 300    & 83.4                   \\
\rowcolor{gray!30}\textbf{ReGLA-X}              & 39.8      & 2.22     & \textbf{3.77} / \textbf{14.81}   & 224   & 300    & \textbf{83.7 (+0.3)}    \\ 
\bottomrule[1.2pt]
\end{tabular}

\label{table-main}
\end{table*}

\subsection{Multi-Teacher Knowledge Distillation} 
To enhance downstream task generalization, we leverage multi-teacher knowledge distillation~\cite{gou2021knowledgedistill} that aggregates complementary expertise from diverse pretrained models. Although many models exhibit strong generalization capabilities, each teacher demonstrates specialized strengths in particular vision tasks despite identical architectures and training data~\cite{asif2020ensembleMT}. Our distillation integrates seven teachers selected for their distinct learning strategies: dBOT-ft~\cite{liu2022dbot} and DeiT-III~\cite{touvron2022deit3} for classification, DINOv2~\cite{oquab2023dinov2} for object detection, SAM2~\cite{ravi2024sam} for semantic segmentation, iBOT~\cite{zhou2021ibot} for masked modeling, ViTamin~\cite{chen2024vitamin} for contrastive learning, and AIMv2~\cite{fini2025multimodal} for autoregressive modeling, comprehensively covering major vision domains.

Building on the UNIC framework~\cite{sariyildiz2024unic}, we introduce key enhancements to optimize knowledge distillation. A ladder encoder with scalable projection heads leverages features from all stages, with the projectors discarded post-distillation to eliminate inference overhead. Student features are resampled to $1/16$ scale for alignment, and teacher outputs undergo feature standardization for consistent ranges. Synchronized batch normalization ensures stable multi-GPU training. The smooth-L1 loss for patch tokens is removed, while other settings align with UNIC. Distillation strategy experiments are in Section 5.4, and teacher model selection is in the supplementary material.

Unlike UNIC, which focuses on classification, our distillation strategy optimizes heterogeneous downstream tasks through two key design choices. First, we build the global class (CLS) token using adaptive average pooling of patch tokens, improving spatial coherence for dense prediction tasks. Second, we omit teacher dropout regularization, as excluding teachers was found to degrade detection and segmentation performance, ensuring consistent knowledge flow from all specialists during training.

\section{Experiments}
\subsection{Image Classification}
We evaluated our models on the ImageNet-1K benchmark~\cite{deng2009imagenet} with $224 \times 224$ inputs and training protocols from recent lightweight architectures~\cite{zheng2025iformer,liu2022convnet}. Most models were trained from scratch for 300 epochs with a 20-epoch linear warm-up to ensure stable optimization. Evaluation includes Top-1 validation accuracy alongside efficiency metrics such as parameter count, FLOPs, and latency. Latency was measured on an iPhone 16 Pro using Core ML Tools with batch size 1, following mobile benchmarking standards~\cite{li2023rethinking,vasu2023mobileone}. Results are reported for base models without distillation (Table~\ref{table-main}) and their distilled versions (Table~\ref{ReGLA-distill}). Full benchmarks are presented in the supplementary materials.

\subsubsection{Results without Multi-Teacher Distillation} 
Table~\ref{table-main} presents a comparative analysis of our models against recent state-of-the-art lightweight models across five distinct parameter scales. Our models consistently achieve the highest Top-1 accuracy within each parameter budget. In particular, ReGLA-T has leading accuracy of 76.4\% in the 3M category. Furthermore, ReGLA-M demonstrates superior performance in the 10M category, achieving 80.85\% accuracy. This represents a significant improvement over MobileNetV4-Conv-M and iFormer-M by 0.95\% and 0.45\%, respectively. In the 20M category, ReGLA-L establishes state-of-the-art performance. Crucially, while utilizing 4.4M fewer parameters than EfficientVit-B2, ReGLA-L delivers 0.7\% higher accuracy.

\subsubsection{Results with Multi-Teacher Distillation} 
As stated in the Methods section, we adopt multi-teacher distillation using seven teacher models. The training regimen comprised an initial 30-epoch distillation phase on ImageNet-21K, followed by 200 epochs of full fine-tuning on ImageNet-1K. The comparison results under distillation, presented in Table~\ref{ReGLA-distill}, demonstrate that our models attain state-of-the-art accuracy within comparable parameter budgets. Notably, ReGLA-M achieves 81.9\% accuracy, exceeding iFormer-M by 0.8\%, while ReGLA-L reaches 84.3\% accuracy, surpassing iFormer-L by 1.6\%. Furthermore, ReGLA-X delivers 0.5\% higher accuracy than FastViT-MA36 while using 3M fewer parameters. Critically, the distillation and fine-tuning strategy consistently yields at least a 1\% accuracy gain over direct training on ImageNet-1K for 200 epochs. These results collectively confirm the efficacy of our proposed distillation strategy.

\begin{table}[htbp]
\centering
\caption{\textbf{Results with distillation.} These models were trained with different teacher models and training strategies.}
\resizebox{0.9\columnwidth}{!}{
\begin{tabular}{c|ccc}

\toprule[1.2pt]
    Model                   & Param. (M)           &Epochs          & Top-1 (\%)      \\ 
\midrule[1pt]
EfficientFormerV2-S1        & 6.1                & 300            & 79.0           \\
MobileViGv2-S              & 7.7                & 300            & 79.8           \\      
FastViT-T12                 & 6.8                & 300            & 80.3           \\
RepViT-M1.1                 & 8.2                & 300            & 80.7           \\ 
iFormer-M                   & 8.9                & 300            & 81.1           \\ 
\rowcolor{gray!30}\textbf{ReGLA-M}      & 9.6                & 230            & \textbf{81.9 (+0.8)} \\ 
\midrule[0.8pt]

SHViT-S4                   & 16.5                & 300            & 80.2           \\
EfficientFormerV2-S2        & 12.6                & 450            & 81.6           \\
MobileViGv2-M               & 15.4                & 300            & 81.7           \\      
RepViT-M1.5                 & 14.0                & 300            & 82.0           \\ 
iFormer-L                   & 14.7                & 300            & 82.7           \\ 
\rowcolor{gray!30}\textbf{ReGLA-L }     & 19.9                & 230            & \textbf{84.3 (+1.6)}  \\ 
\midrule[0.8pt]

FastViT-SA36                & 30.4                & 300            & 84.2           \\
FastViT-MA36                & 42.7                & 300            & 84.6           \\
\rowcolor{gray!30}\textbf{ReGLA-X }     & 39.8                & 230            & \textbf{85.2 (+0.6)}  \\ 
\bottomrule[1.2pt]

\end{tabular}
}
\label{ReGLA-distill}
\end{table}

\begin{figure}[h]
\centering
\includegraphics[width=1.0\columnwidth]{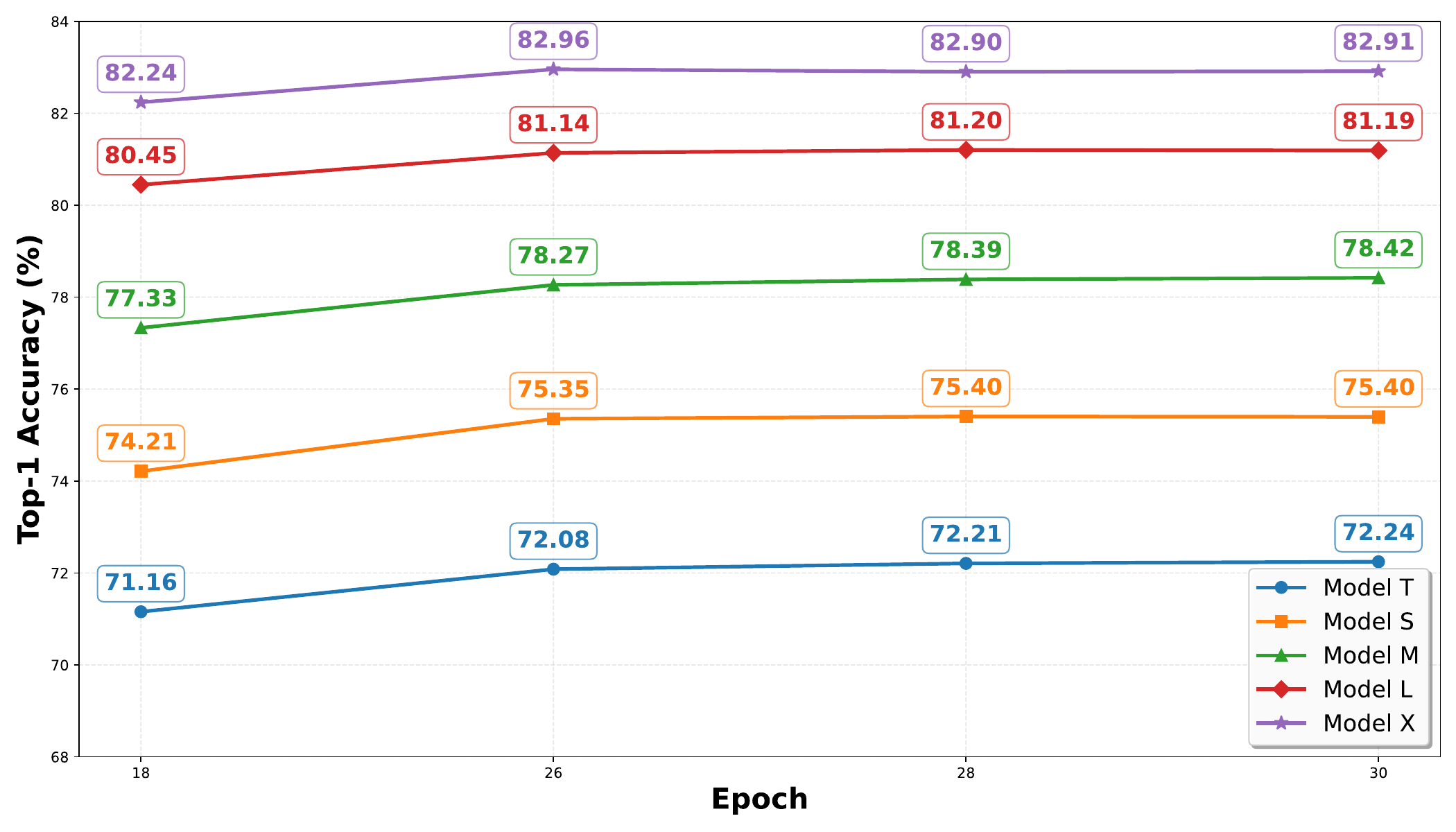} % Reduce the figure size so that it is slightly narrower than the column. Don't use precise values for figure width.This setup will avoid overfull boxes.
\caption{Top-1 accuracy progression during distillation, measured via KNN (n=20) classification, shows diminishing improvements after the 18th epoch and only marginal gains up to the 30th. Training for 30 epochs strikes an optimal balance between computational cost and performance.}
\label{fig_distill_epoch}
\end{figure}

\textbf{Distillation and Generalization Evaluation.} During the distillation phase, training was limited to 30 epochs due to diminishing returns observed with additional training. As shown in Fig.~\ref{fig_distill_epoch}, the Top-1 accuracy, measured using KNN (n=20) classification, shows a noticeable slowdown in improvement after the 18th epoch, indicating minimal benefits from further training. By carefully balancing computational cost and accuracy, we determined that 30 epochs achieve an optimal trade-off, ensuring efficient resource utilization while delivering high-quality distillation results.

% \begin{figure}[h]
% \centering
% \includesvg[width=1.0\columnwidth]{Image/distill_epoch11.svg}
% \caption{Top-1 accuracy progression during distillation, measured via KNN (n=20) classification, shows diminishing improvements after the 18th epoch and only marginal gains up to the 30th. Training for 30 epochs strikes an optimal balance between computational cost and performance.}
% \label{fig_distill_epoch}
% \end{figure}

\begin{table*}[!t]
\centering
\caption{\textbf{Object Detection and Instance Segmentation}  results on COCO2017 using Mask RCNN. \textbf{Semantic Segmentation} results on ADE20K using the Semantic FPN framework. We measure all backbone latencies with image crops of $512 \times 512$ on iPhone16 Pro by CoreML Tools. Failed indicates that the model runs too long to report latency by the Core ML.}
% \resizebox{0.95\columnwidth}{!}{
\begin{tabular}{c|c|c|ccc|ccc|c}  % 列格式：左对齐+右对齐，竖线分隔
\toprule[1.2pt]
\multicolumn{1}{c|}{ \multirow{2}{*}{Backbone} } & \multicolumn{1}{c|}{Param.}  
& \multicolumn{1}{c|}{Latency}
& \multicolumn{3}{c|}{Object Detection} 
& \multicolumn{3}{c|}{Instance Segmentation} & \multicolumn{1}{c}{Semantic} \\ \cline{3-5} \cline{6-8} \cline{9-9}  % 子列分隔线
& \multicolumn{1}{c|}{(M)}
& \multicolumn{1}{c|}{(ms)}
& \multicolumn{1}{c}{AP$^\text{box}$} & \multicolumn{1}{c}{AP$_{50}^\text{box}$} & \multicolumn{1}{c|}{AP$_{75}^\text{box}$}
& \multicolumn{1}{c}{AP$^\text{mask}$} & \multicolumn{1}{c}{AP$_{50}^\text{mask}$} & \multicolumn{1}{c|}{AP$_{75}^\text{mask}$} 
& \multicolumn{1}{c}{mIoU} \\ 
\midrule[0.8pt]

EfficientNet-B0~\cite{tan2019efficientnet}          & 5.3   & 8.89    & 31.9  & 51.0  & 34.5  & 29.4  & 47.9  & 31.2  & -     \\
PoolFormer-S12~\cite{yu2022metaformer}              & 11.9  & 9.22    & 37.3  & 59.0  & 40.1  & 34.6  & 55.8  & 36.9  & 37.2  \\
EfficientFormer-L1~\cite{li2022efficientformer}     & 12.3  & 10.84    & 37.9  & 60.3  & 41.0  & 35.4  & 57.3  & 37.3  & 38.9  \\
FastViT-SA12~\cite{vasu2023fastvit}                 & 10.9  & 7.35    & 38.9  & 60.5  & 42.2  & 35.9  & 57.6  & 38.1  & 38.0  \\
RepViT-M1.1~\cite{wang2023repvit}                   & 8.2   & 8.30    & 39.8  & 61.9  & 43.5  & 37.2  & 58.8  & 40.1  & 40.6  \\
CAS-ViT-M~\cite{zhang2024cas}                       & 12.4  & 9.23    & 41.2  & 63.8  & 44.7  & 38.4  & 61.3  & 40.9  & 43.1  \\
iFormer-M-distlled~\cite{zheng2025iformer}          & 8.9   & 19.24    & 40.8  & 62.5   & 44.8  & 37.9  & 59.7  & 40.7  & 42.4  \\   
\midrule[0.8pt]

\rowcolor{gray!30}\textbf{ReGLA-M}                              & 9.6   & 4.98        & \textbf{41.7}  & \textbf{64.3}  & \textbf{45.6}  
                                                            & \textbf{38.6}  & \textbf{61.5}  & \textbf{41.4}  & \textbf{43.3}  \\   
\rowcolor{gray!30}\textbf{ReGLA-M-distilled}                      & 9.6   & 4.98        & \textbf{43.9}  & \textbf{65.8}  & \textbf{48.3}  
                                                            & \textbf{40.4}  & \textbf{63.0}  & \textbf{43.2}  & \textbf{46.0}  \\  
\midrule[0.8pt]

PoolFormer-S24~\cite{yu2022metaformer}              & 21.4    & 16.10    & 40.1  & 62.2  & 43.4  & 37.0  & 59.1  & 39.6  & 40.3  \\
ConvNeXt-T~\cite{liu2022convnet}                    & 29.0    & 13.06    & 41.0  & 62.1  & 45.3  & 37.7  & 59.3  & 40.4  & 41.4  \\
EfficientFormer-L3~\cite{li2022efficientformer}     & 31.3    & 18.56    & 41.4  & 63.9  & 44.7  & 38.1  & 61.0  & 40.4  & 43.5  \\
RepViT-M1.5~\cite{wang2023repvit}                   & 14.0    & 13.55    & 41.6  & 63.2  & 45.3  & 38.6  & 60.5  & 41.5  & 43.6  \\
PVTv2-B1~\cite{wang2022pvt}                         & 14.0    & 19.25    & 41.8  & 64.3  & 45.9  & 38.8  & 61.2  & 41.6  & 42.5  \\
FastViT-SA24~\cite{vasu2023fastvit}                 & 20.6    & 13.25    & 42.0  & 63.5  & 45.8  & 38.0  & 60.5  & 40.5  & 41.0  \\
EfficientMod-S~\cite{ma2024efficient}               & 32.6    & 16.5     & 42.1  & 63.6  & 45.9  & 38.5  & 60.8  & 41.2  & 43.5  \\
Swin-T~\cite{liu2021swin}                           & 28.3    & Failed   & 42.2  & 64.4  & 46.2  & 39.1  & 61.6  & 42.0  & 41.5  \\
CAS-ViT-T~\cite{zhang2024cas}                       & 21.8    & 15.12    & 43.5  & 65.3  & 47.5  & 39.6  & 62.3  & 42.2  & 45.0  \\
iFormer-L-distlled~\cite{zheng2025iformer}            & 14.7    & 31.10    & 42.2  & 64.2  & 46.0  & 39.1  & 61.4  & 41.9  & 44.5  \\  
\midrule[0.8pt]

\rowcolor{gray!30}\textbf{ReGLA-L}                              & 19.9    & 8.82    & \textbf{44.2}  & \textbf{66.5}  & \textbf{48.4}  
                                                            & \textbf{40.3}  & \textbf{63.3}  & \textbf{43.3}  & \textbf{45.4}  \\ 
\rowcolor{gray!30}\textbf{ReGLA-L-distilled}                      & 19.9   & 8.82      & \textbf{46.2}  & \textbf{67.9}  & \textbf{50.6}  
                                                          & \textbf{42.1}  & \textbf{65.1}  & \textbf{45.8}  & \textbf{48.4}  \\ 
\bottomrule[1.2pt]

\end{tabular}
\label{table:coco_ade}
\end{table*}

\begin{figure}[h]
\centering
\includegraphics[width=1.0\columnwidth]{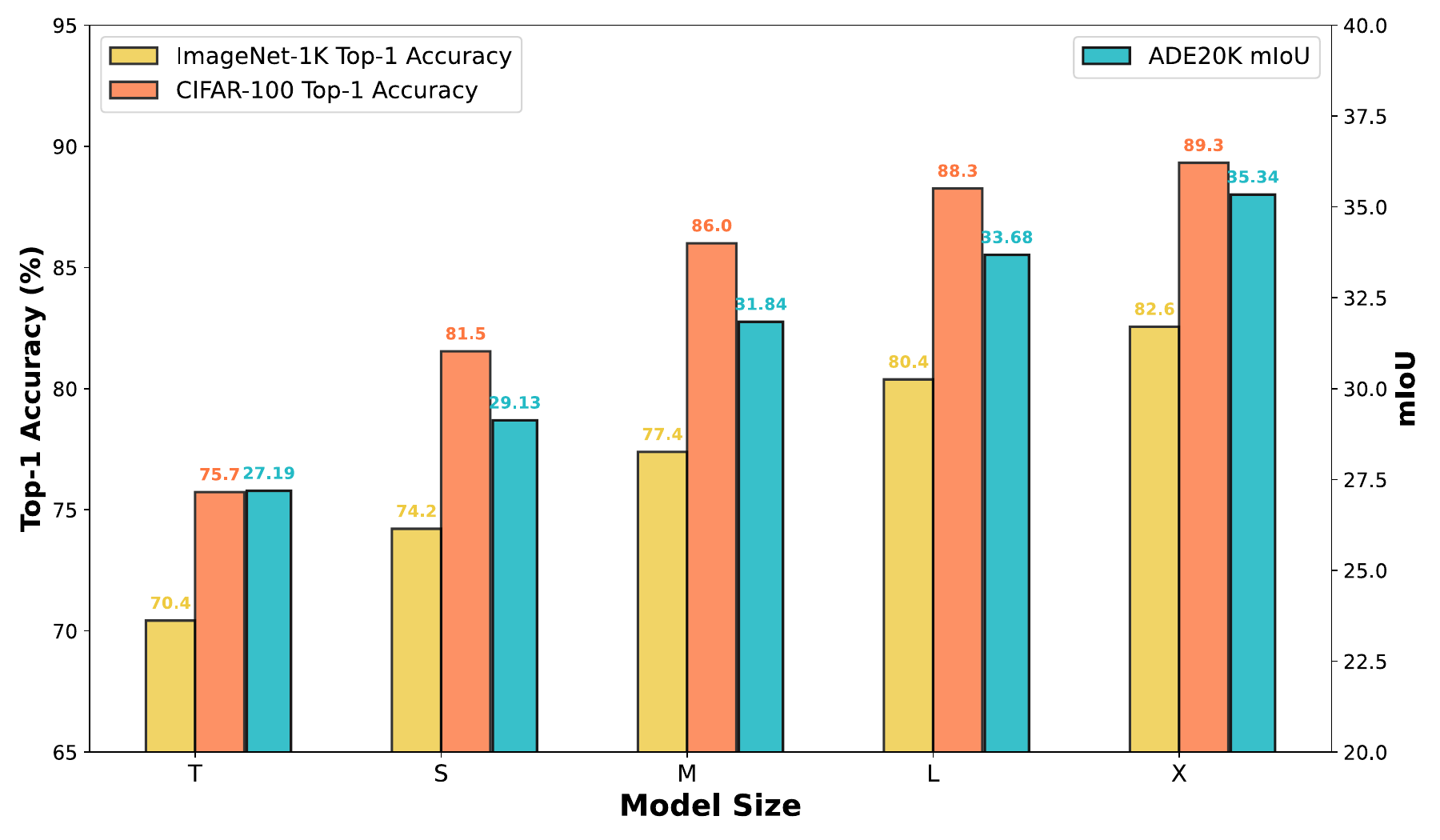} % Reduce the figure size so that it is slightly narrower than the column. Don't use precise values for figure width.This setup will avoid overfull boxes.
\caption{Performance comparison across various model sizes on multiple benchmarks. Left: Top-1 accuracy (\%) measured using KNN (n=20) classification on ImageNet1K and CIFAR100 steadily ascends with increasing model size. Right: Mean Intersection over Union (mIoU) on ADE20K also rises as increasing model size, demonstrating strong scalability and generalization across tasks.}
\label{fig_cifer}
\end{figure}

We conducted 30 epochs of distillation on ImageNet-21K using our models, followed by KNN (n=20) classification experiments on ImageNet-1K and CIFAR100, and segmentation performance evaluation on ADE20K. The results in Fig.~\ref{fig_cifer} demonstrate that our models do not suffer from overfitting and exhibit excellent scaling properties, making them capable of scaling to models with larger parameter counts.

% \begin{figure}[h]
% \centering
% \includesvg[width=1.0\columnwidth]{Image/distill_eval_bar.svg}
% \caption{Performance comparison across various model sizes on multiple benchmarks. Left: Top-1 accuracy (\%) measured using KNN (n=20) classification on ImageNet-1K and CIFAR100 steadily improves with increasing model size. Right: Mean Intersection over Union (mIoU) on ADE20K also rises as model size scales, demonstrating strong scalability and generalization across tasks.}
% \label{fig_cifer}
% \end{figure}

\subsection{Transfer Learning}
\subsubsection{Object Detection and Instance Segmentation}
To validate the transferability of ReGLA architectures, we conducted comprehensive object detection and instance segmentation experiments using Mask R-CNN~\cite{he2017maskrcnn} within the MMDetection framework~\cite{chen2019mmdetection}. Our implementation followed iFormer's experimental protocols~\cite{zheng2025iformer}, training all models for 12 epochs with identically configured parameters. We evaluated both non-distilled and distilled variants initialized with corresponding ImageNet pretrained weights.

As summarized in Table~\ref{table:coco_ade}, ReGLA consistently achieves state-of-the-art performance on both AP$^\text{box}$ and AP$^\text{mask}$ metrics. The non-distilled ReGLA-M surpasses iFormer-M-distill by nearly 1\% in both tasks. When employing distillation, ReGLA-M-distilled exhibits more substantial gains, outperforming iFormer-M-distilled by +3.1\% AP$^\text{box}$ and +2.5\% AP$^\text{mask}$. Remarkably, ReGLA-M-distilled even exceeds the large-scale iFormer-L-distilled by +1.7\% AP$^\text{box}$ and +1.3\% AP$^\text{mask}$. For large-scale models, ReGLA-L demonstrates a 2\% AP$^\text{box}$ gain over iFormer-L-distilled in object detection, while its distilled counterpart achieves a more significant 4\% improvement. Similarly, in instance segmentation, ReGLA-L and ReGLA-L-distilled outperform iFormer-L-distilled by 1.2\% and 3.0\% AP$^\text{mask}$, respectively. These consistent performance improvements highlight the exceptional downstream transfer capability of our architectures.

\subsubsection{Semantic Segmentation}
In addition, we evaluated segmentation capability of ReGLA on ADE20K~\cite{zhou2017ade20k} using Semantic FPN~\cite{kirillov2019panoptic} within the MMSegmentation framework~\cite{contributors2020mmsegmentation}. Following established practices in~\cite{li2022efficientformer}, our training configuration mirrored the protocol: models underwent 80,000 iterations with a fixed learning rate of 0.0001. We evaluated both non-distilled and distilled variants of ReGLA-M and ReGLA-L.

As reported in Table~\ref{table:coco_ade}, ReGLA achieves superior segmentation performance measured by mIoU. The distilled ReGLA-M exhibits a 3.6\% mIoU advantage over its iFormer-M-Distilled counterpart. More significantly, ReGLA-M-Distilled surpasses the larger iFormer-L-Distilled by 1.5\% mIoU, demonstrating exceptional scalability. For large-scale models, ReGLA-L-Distilled further extends this advantage with a 3.9\% mIoU improvement compared to iFormer-L-Distilled. These consistent performance advantages across model scales confirm the effectiveness of the architecture for dense prediction tasks.

\section{Ablation Studies}
We conducted ablation studies on ReGLA-M to validate the design choices of its core components. Unless otherwise specified, all models were trained under the same protocol on ImageNet-1K and evaluated by Top-1 accuracy. In the tables below, Conv.k denotes standard depth-wise convolutions with kernel sizes of $k\times k$. Additional ablations are provided in the supplementary materials.

\subsection{Early-stage Feature Extraction}
We investigated the efficiency of the ELRF module by replacing it with $7\times7$ and $9\times9$ convolution kernels in early stages. Table~\ref{ReGLA-ablation-ELRF} shows $9\times9$ kernels provide minimal gains over $7\times7$. In contrast, the ELRF module significantly improves Top-1 accuracy by capturing local information and maintaining a large receptive field, enhancing overall performance.

\begin{table}[htbp]
\centering
\caption{Impact of convolution kernel sizes and the ELRF module on performance. The ELRF module significantly improves Top-1 accuracy by capturing local information while maintaining a large receptive field.}
\resizebox{0.95\columnwidth}{!}{
\begin{tabular}{c|ccc|c}
\toprule[1.2pt]
 & Stages 1 \& 2    & Attention     & After Attn.    & Top-1 (\%) \\ \midrule[0.8pt]
1& Conv.9         & ReLU (gate*)   & Conv.3         & 80.49 \\
2& Conv.7         & ReLU (gate*)   & Conv.3         & 80.44 \\
3& ELRF          & ReLU (gate*)   & Conv.3         & 80.65 \\ 
\bottomrule[1.2pt]
\end{tabular}
}
\label{ReGLA-ablation-ELRF}
\end{table}

\subsection{The Gated Mechanism in Attention}
Table~\ref{ReGLA-ablation-attn} reports the ablation results for the attention module. The notation gate* indicates that the value branch (V) computation is decoupled from the gated branch in the gated mechanism. Incorporating a gated mechanism into the ReLU linear attention improves Top-1 accuracy by 0.19\%, while integrating the value branch with the gated modulation further boosts accuracy by 0.2\%, achieving the best performance in this setup.

\begin{table}[htbp]
\centering
\caption{Ablation of gating strategies in attention. Full integration of the gating mechanism with the value branch delivers the best performance.}
\resizebox{0.95\columnwidth}{!}{
\begin{tabular}{c|ccc|c}
\toprule[1pt]
 & Stages 1 \& 2    & Attention     & After Attn.    & Top-1 (\%) \\ 
\midrule[0.8pt]
1& ELRF        & ReLU (no gate) & Conv.3         & 80.46 \\
2& ELRF        & ReLU (gate*)   & Conv.3         & 80.65 \\
\rowcolor{gray!30}3& ELRF        & ReLU (gate)    & Conv.3         & \textbf{80.85} \\ 
\bottomrule[1pt]
\end{tabular}
}
\label{ReGLA-ablation-attn}
\end{table}

\subsection{Post-Attention Module Selection}
Following the attention block, we compared three common design choices: a standard FFN, a MobileNetV2-style inverted bottleneck (MIB)~\cite{sandler2018mobilenetv2}, and a lightweight $3\times3$ convolution (Conv3). Surprisingly, \textbf{the simplest option (i.e., Conv3) achieves the highest accuracy (80.85\%)}, outperforming both FFN (80.65\%) and MIB (80.70\%), as shown in Table~\ref{ReGLA-ablation-afterattn}.

\begin{table}[htbp]
\centering
\caption{Comparison of post-attention modules. A lightweight $3\times3$ convolution strikes the optimal balance between expressiveness and efficiency.}
\resizebox{0.95\columnwidth}{!}{
\begin{tabular}{c|ccc|c}
\toprule[1.2pt]
 &Stages 1 \& 2 & Attention & After Attn. & Top-1 (\%) \\ 
\midrule[0.8pt]
1& ELRF & ReLU (gate) & FFN & 80.65 \\
2& ELRF & ReLU (gate) & MIB & 80.70 \\
\rowcolor{gray!30}3& ELRF & ReLU (gate) & Conv.3 & \textbf{80.85} \\ 
\bottomrule[1.2pt]
\end{tabular}
}
\label{ReGLA-ablation-afterattn}
\end{table}

\subsection{Distillation Strategy Optimization}
We conducted distillation experiments on ImageNet-1K, training for 100 epochs and using KNN (n=20) to calculate Top-1 accuracy. As shown in Table~\ref{Multi-teacher-exps}, we started with the UNIC framework, but our model initially used only patch tokens, as it lacked a CLS token. After introducing the CLS token, the model's performance improves by 4.5\%. Subsequently, we incorporated additional distillation techniques, such as synchronized batch normalization and using only cosine similarity loss, to further enhance performance. After finalizing our distillation strategy (as seen in the 4-th row of Table~\ref{Multi-teacher-exps}), we progressively added more teacher models until a total of seven were utilized.

\begin{table}[htbp]
\centering
\caption{\textbf{Selection of teacher models and distillation strategies.} \textbf{4tea} indicates the use of four teacher models: dBoT-ft, DeiT-III, DINOv2, and iBoT; \textbf{6tea} adds ViTamin and AIMv2 to the 4tea setup, and \textbf{7tea} further includes SAM2. \textbf{UNIC} represents the full adoption of the UNIC distillation strategy except CLS token, \textbf{addycls} denotes we manually add the CLS token, \textbf{syBN} denotes the use of synchronized batch normalization, and \textbf{onlycos} signifies only use cosine similarity loss.}
\resizebox{0.95\columnwidth}{!}{
\begin{tabular}{c|c|l|c}
\toprule[1.2pt]
&Teacher  & Distillation Strategy                       & Top-1 (\%) \\ 
\midrule[0.8pt]
1 & 4tea           & UNIC                               & 72.662 \\
2 & 4tea           & UNIC + onlycos                     & 74.326 \\ 
3 & 4tea           & UNIC + onlycos +  addcls           & 77.084 \\
\rowcolor{gray!30}4 & 4tea           & UNIC + onlycos +  addcls + syBN    & 77.66 \\
5 & 6tea           & UNIC + onlycos +  addcls + syBN    & 78.05 \\ 
\rowcolor{gray!30}6 & 7tea           & UNIC + onlycos +  addcls + syBN    & 78.424 \\ 
\bottomrule[1.2pt]
\end{tabular}
}
\label{Multi-teacher-exps}
\end{table}

\section{Conclusion}
We present ReGLA, a lightweight hybrid architecture designed for high-resolution images. By combining local and global modeling, ReGLA balances accuracy with on-device latency. The ELRF module captures fine-grained spatial details, while the RGMA module enables stable, expressive global modeling with linear complexity. Experiments show that global attention can be efficient: RGMA surpasses softmax-based methods in both speed and accuracy under resource constraints. Additionally, co-designing local operators and attention mechanisms is critical for high-resolution efficiency. Distilled variants achieve state-of-the-art performance via a refined multi-teacher strategy in downstream tasks.

{
    \small
    \bibliographystyle{ieeenat_fullname}
    \bibliography{main}
}

% WARNING: do not forget to delete the supplementary pages from your submission 
% \input{sec/X_suppl}

\end{document}